\documentclass[letterpaper]{article} 
\usepackage{aaai25}
\usepackage{times}  
\usepackage{helvet}  
\usepackage{courier}  
\usepackage[hyphens]{url}  
\usepackage{graphicx} 
\usepackage{natbib}  
\usepackage{caption} 
\usepackage{algorithm}
\usepackage{algorithmic}

\usepackage{newfloat}
\usepackage{listings}
\DeclareCaptionStyle{ruled}{labelfont=normalfont,labelsep=colon,strut=off} 
\floatstyle{ruled}
\newfloat{listing}{tb}{lst}{}
\floatname{listing}{Listing}
\usepackage{tikz}
\usepackage{amsmath}
\usepackage{multirow}
\usepackage{makecell}
\usepackage{booktabs}    
\usepackage{colortbl}
\newcommand{\nn}{\nonumber \\}
\newcounter{chatlinenum}

\tikzset{chatstyle/.style={text width=2.2in,rounded corners=2pt}}

\definecolor{mygreen}{HTML}{88EABB}
\definecolor{myblue}{HTML}{80bdff}
\definecolor{mybluee}{HTML}{1987ff}
\definecolor{myred}{HTML}{ed2939}

\def\chatline#1\par{%
   \stepcounter{chatlinenum}%
   \noindent
   \ifodd\thechatlinenum
       \tikz[]{\node[fill=lightgray,chatstyle]{\strut#1\strut};}%
   \else
       \tikz[]{\node[fill=mygreen,chatstyle]{\strut#1\strut};}%
   \fi
   \par
   \smallskip
}

\begingroup
    \lccode`~=`\^^M
    \lowercase{%
\endgroup
    \def\newchatline#1~{%
        \stepcounter{chatlinenum}%
        \ifodd\thechatlinenum
            \tikz[]{\node[fill=lightgray,chatstyle]{\strut#1\strut};}%
        \else
            \hfill
            \tikz[]{\node[fill=myblue,chatstyle,align=left]{\strut#1\strut};}%
        \fi
        ~
        \smallskip
    }%
}

\NewDocumentEnvironment{newchat}{}{%
    \setcounter{chatlinenum}{1}
    \begin{minipage}{\columnwidth}
        \obeylines
        \everypar={\newchatline}
}{%
    \end{minipage}
}

\newcommand{\smalltonormalsize}{%
  \fontsize
    {\fpeval{(\f@size@small+\f@size@normalsize)/2}}
    {\fpeval{(\f@baselineskip@small+\f@baselineskip@normalsize)/2}}%
  \selectfont
}

\title{Hansel: Output Length Controlling Framework for Large Language Models}
\author{
    Seoha Song\textsuperscript{\rm 1}\equalcontrib, 
    Junhyun Lee\textsuperscript{\rm 1}\equalcontrib\thanks{Corresponding author.}, 
    Hyeonmok Ko\textsuperscript{\rm 1}
}
\affiliations{
    \textsuperscript{\rm 1}Samsung Research\\

    \{seoha.song, junhyun8.lee, felix.ko\}@samsung.com
}

\begin{document}

\maketitle

\begin{abstract}
Despite the great success of large language models (LLMs), efficiently controlling the length of the output sequence still remains a challenge. 
In this paper, we propose Hansel, an efficient framework for length control in LLMs without affecting its generation ability. 
Hansel utilizes periodically outputted hidden special tokens to keep track of the remaining target length of the output sequence.
Together with techniques to avoid abrupt termination of the output, this seemingly simple method proved to be efficient and versatile, while not harming the coherency and fluency of the generated text.
The framework can be applied to any pre-trained LLMs during the finetuning stage of the model, regardless of its original positional encoding method. 
We demonstrate this by finetuning four different LLMs with Hansel and show that the mean absolute error of the output sequence decreases significantly in every model and dataset compared to the prompt-based length control finetuning. 
Moreover, the framework showed a substantially improved ability to extrapolate to target lengths unseen during finetuning, such as long dialog responses or extremely short summaries. 
This indicates that the model learns the general means of length control, rather than learning to match output lengths to those seen during training.
\end{abstract}

%

\section{Introduction}

\begin{figure}[ht]
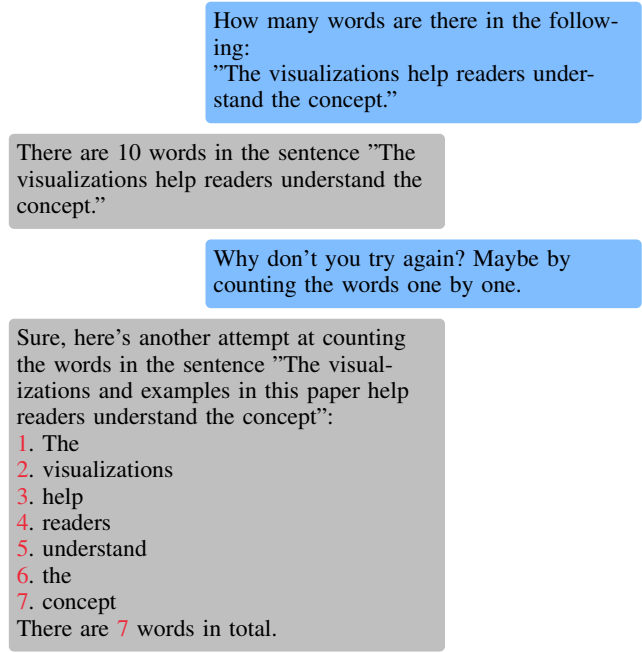

    \centering
    \small
    \begin{newchat}
        How many words are there in the following: \\"The visualizations help readers understand the concept." 
        There are 10 words in the sentence "The visualizations help readers understand the concept."
        Why don't you try again? Maybe by counting the words one by one.
        Sure, here's another attempt at counting the words in the sentence "The visualizations and examples in this paper help readers understand the concept": \\ {\color{myred}1}. The\\{\color{myred}2}. visualizations\\{\color{myred}3}. help\\{\color{myred}4}. readers\\{\color{myred}5}. understand\\{\color{myred}6}. the\\{\color{myred}7}. concept\\There are {\color{myred}7} words in total.    
    \end{newchat}
    \caption{An example conversation with GPT.
    We have edited the example sentence for brevity.}
    \label{fig:example_chat}
\end{figure}

Large language models (LLMs) have shown great promise in generating coherent and fluent text, making them useful for a variety of tasks~\cite{arivazhagan2019massively, karpukhin2020dense, lewis2020bart, zhang2020pegasus}. 
However, a number of challenges still remain for LLMs, including the difficulty of efficiently controlling the length of the output sequence. 
Length control of the output sequence has practical importance in the real-world application of LLMs. 
Setting the length determines the level of details to include in the output, which is desired both in the aspect of content and system interface.
While this task seems relatively mundane compared to the emergent abilities of LLMs, to the best of our knowledge length control has yet to be achieved even for the larger LLMs~\cite{achiam2023gpt, anil2023palm, touvron2023llama} with hundreds of billions of parameters.

A couple of specific examples where length control is useful would be i) a news app where articles are accompanied by AI-generated summaries, or ii) a voice assistant that can tune the amount of information to be spoken.  
For the news app, the default length should be concise enough to fit the system interface, and the user may want to tune the summarization length to include or drop details. 
Even for simple voice assistant tasks such as telling the weather, ``rain at noon'' may be sufficient but sometimes precipitation or temperature might be needed. 
These examples show the benefits of various outputs from the identical input and controlling them is important and useful. 

The study of length control dates back to the pre-transformer era.
\citet{kikuchi2016controlling} first suggested feeding the remaining length information to the LSTM as an embedding or the memory cell. 
After the advent of the transformer~\cite{vaswani2017attention}, there were attempts to include the length information in a modified form of positional encoding~\cite{takase-okazaki-2019-positional}.
Other studies turn to the attention mechanism such as Length Attention~\cite{yu-etal-2021-lenatten} and Length-Aware Attention~\cite{liu2022length}.

Although the previous methods of length control were fairly successful, they mostly concentrate on the relatively small sequence-to-sequence models~\cite{sutskever2014sequence} and not the nowadays prevalent decoder models.
The main reason for this is two-fold.
First, the aforementioned methods require application in the pretraining stage and show significant performance degradation when used only while finetuning. 
Large decoder models are mostly pre-trained once as a foundation model and finetuned for downstream tasks due to their large size, making the previous methods inapplicable.
Second, with their better understanding of the context, the decoder models have some, although imperfect, length-controlling ability. 
The decoder LLMs would not output the exact required length, for example, 17 words, but it does understand commands such as shorter/longer and will tend to output longer texts when specified to output 25 words than when asked for 10 words.

In this paper, we introduce a simple \emph{finetuning} framework to improve the output length-controlling ability of LLMs. 
We first observe why generating outputs with a certain length is a difficult task for LLMs.
LLMs are known to have limited mathematical ability even with the significant improvement from the zero-shot-CoT technique~\cite{kojima2022large}.
This not only applies to complex math problems but also to simple counting. 
Figure~\ref{fig:example_chat} shows a short example conversation with GPT~\footnote{
We have used GPT-3.5 to generate the example. GPT-4 can generate similar conversations by explicitly instructing not to call Python scripts.
}, slightly edited for brevity.
Because LLMs are not intrinsic counters, GPT sometimes outputs incorrect answers when asked to answer the number of words in a sentence.
One way to correct this behavior is to prompt the model to explicitly count the words first and then output the answer as in Figure~\ref{fig:example_chat}. 

We build upon this observation and apply it to the task of output length control. 
With its lack of counting ability, the model would not be aware of how many words it has output and how many words are left at a certain point.
To overcome this, we introduce the Hansel -- \textbf{H}idden \textbf{A}rrangements (of special tokens) in \textbf{N}atural \textbf{S}equence for \textbf{E}xpected \textbf{L}ength --  framework, which directly injects the information via the form of hidden special tokens. 
By augmenting a finetuning dataset with special tokens appearing regularly, the model can efficiently learn the relative position compared to the target length at a given time. 
This can be considered a simple hard-coded CoT method specifically for output length control. 
Our method performs well even when only used during finetuning without  degradation in output quality. 
Phi2~\cite{li2023textbooks} based Hansel model showed great performance in mean absolute error from the reference length in four summarization and dialogue datasets (Table~\ref{tab:phi2}).  
Moreover, the error remains small for a wide range of arbitrary target lengths, where other methods diverge rapidly as the target length deviates from the dataset's typical length (Figure~\ref{fig:extrapolation}).

\paragraph{Contribution} The main contributions of this paper are as follows:
\begin{itemize}
    \item We introduce the Hansel framework that learns the output length control during finetuning through repeated hidden tokens, which can be applied to decoder LLMs with large number of parameters.
    \item {We empirically validate the effectiveness of Hansel by experimenting with it on four different datasets, various models, and several target lengths.} 
    \item {We conduct a thorough comparison with the prompt-based length control method and show that Hansel outperforms the prompt-based method as it learns the general means of length control. }
\end{itemize}

\section{Related Work}
\label{sec:related}

\begin{figure*}[t]
    \centering
    \includegraphics[width=0.79 \textwidth]{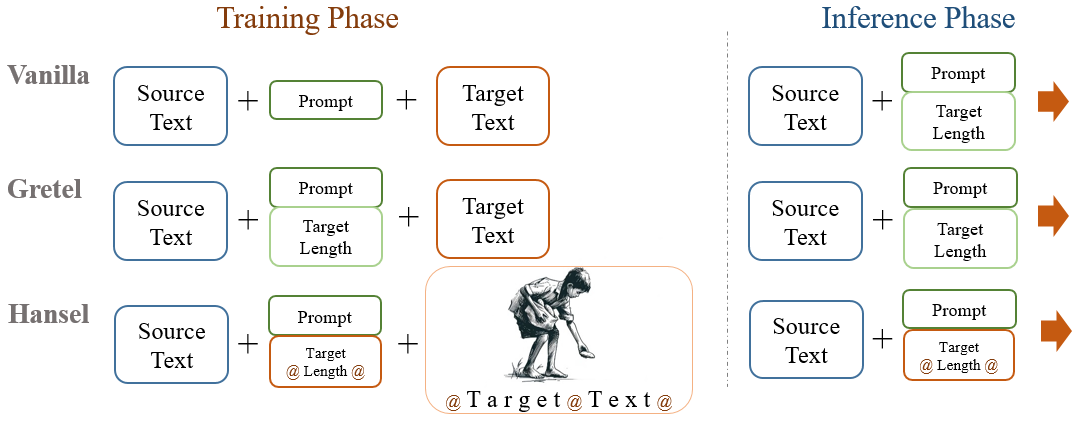}
    \caption{Schematic of the Hansel framework, compared with the vanilla and Gretel scheme. Vanilla is normal fine-tuning and Gretel is the prompt-based length-aware fine-tuning. 
    Hansel receives the target length as a special token and regularly places additional special tokens (marked as @ in the figure) that inform the position while fine-tuning. 
    } 
    \label{fig:hansel}
\end{figure*}

Majority of previous studies have been done with sequence-to-sequence models. 
LenEmb and LenInit~\cite{kikuchi2016controlling} are learning-based methods where the length information is provided via an extra embedding and the LSTM memory cell, respectively.
\citet{takase-okazaki-2019-positional} modified the Transformer's positional encoding to include the remaining length information. 
\citet{fan2018controllable} train on examples with the length information (in ranges) included as a special marker and use the target length marker during inference. 
Other methods include utilizing the attention mechanism~\cite{yu-etal-2021-lenatten, liu2022length} and prompts~\cite{zhang2022latent, he-etal-2022-ctrlsum}.
While showing some promise in length-controlling ability, most methods previously mentioned require special pre-training and/or structural modification to the baseline model.
Our method has some similarities with \citet{fan2018controllable} in the sense that we use a special marker, and \citet{zhang2022latent} in that we utilize prompts.
However, we use the special tokens periodically which is the crucial difference.

Decoder LLMs are now more common than encoder-decoder models, and they mostly come in more than billions of parameters.
They tend to intrinsically have some degree of sense of the output length, and may properly reply to prompts such as ``give me a shorter/longer answer.'' 
The paradigm of length control in such models involves prompt learning~\cite{raffel2020exploring, sanh2022multitask, 10.1145/3560815}. 
\citet{goyal2022news} showed good results on prompt-based length control but only in the units of sentences.
There is also work on prompt-based length control in the framework of reinforcement learning~\cite{jie2023prompt}.

Recently, a new set of studies on length-control have emerged in the context of direct preference optimization (DPO)~\cite{rafailov2024direct}. 
DPO tends to generate longer responses~\cite{singhal2024a}, which is in part because they are preferred over shorter ones while evaluation~\cite{dubois2024alpacafarm}.
To prevent this over-verbosity, a number of papers have suggested methods to length-control, or length-desensitize during DPO. 
These include R-DPO~\cite{park2024disentangling}, SimPO~\cite{meng2024simpo}, LIFT-DPO~\cite{yuan2024following}, SamPO~\cite{lu2024eliminating}, and LD-DPO~\cite{liu2024length}.
However, with a different motivation, these works do not require outputs of specific lengths.

\section{Method}
\label{sec:hansel}

As we have observed in Figure~\ref{fig:example_chat}, LLMs are intrinsically not well suited for counting. 
The key idea of Hansel is to overcome such lack of counting ability by explicit counting and use it for counting the output words.
In Hansel, we augment the fine-tuning training set by placing special tokens that indicate the remaining number of words.
When fine-tuned with such dataset, the special tokens will also appear while inference and guide the model to output the desired length.

\subsection{The Hansel dataset}

The Hansel framework utilize the ``Hansel dataset,'' which is the original dataset augmented with special tokens that indicate the remaining length of the output. (Thus, giving the name Hansel from the German fairy tale ``Hansel and Gretel.'' Hansel leaves a trail of pebbles in the woods to find the way back home.)
From now on, we count the units of length as \emph{words} unless specified. 
The method is trivial to apply in different units, such as characters, tokens, and sentences.
Empirically, the difficulty of length control increases from sentence, token, word, and character. 

Below we take an example from CNN/DM.
\begin{align}
&[\textrm{Original}]: \nn
&\textit{Famous American foods created across United States. } \nn 
&\textit{Connecticut diner claims creation of the hamburger.} \nn
&\textit{Onion rings were courtesy of cook at Pig Stand in Texas.} \nn[4pt]
&[\textrm{Hansel}]: \nn
&{{|2\rangle \langle 5|}} \textit{Famous American foods created across } {|2\rangle} \textit{United} \nn 
&\textit{States. Connecticut diner claims creation of the hambur-} \nn
&\textit{ger. Onion } {|1\rangle} \textit{rings were courtesy of cook at Pig Stand} \nn
&\textit{in Texas. }{|0\rangle} \nonumber 
\end{align}
%
The augmented special token $|x\rangle \langle y|$ means that there are $\Delta x + y$ words left until the target length and $|x\rangle \equiv |x\rangle \langle 0|$.
Hyperparameter $\Delta$ is the stride between special tokens, excluding the very first special token. 

For each example, we count the number of words in the reference output ($l$) and include $| \lfloor{l/\Delta} \rfloor \rangle \langle l \% \Delta|$ at the beginning. After $l\%\Delta$ words we insert $| \lfloor{l/\Delta} \rfloor \rangle $. The special tokens $| \lfloor{l/\Delta}\rfloor - 1 \rangle, | \lfloor{l/\Delta}\rfloor - 2 \rangle, \cdots, | 0 \rangle $ are inserted in $\Delta$ words interval until the end of the reference. 
{In the above CNN/DM example, $l = 25$ and $\Delta = 10$. 
Since $\lfloor{l/\Delta} \rfloor = 2$ and $l\%\Delta = 5$, the first special token becomes $|2\rangle \langle5|$ and $|2\rangle$ follows after 5 words.}

When finetuning with the above dataset, there is a risk that the output will be abruptly terminated after the $| 0 \rangle$ token without actually finishing the sentence.
To avoid this, we introduce one more hyperparameter $\delta$, which indicates the maximum residual words after the $| 0 \rangle$ token.
We set aside a portion of examples and consider the total length as $l-1, l-2, \cdots, l-\delta$ instead of $l$.
The purpose of a nonzero $\delta$ is to train with examples where the response does not terminate with $| 0 \rangle$, and thus educate the model to finish the sentence naturally even when an unfinished response encounters $| 0 \rangle$.
The previous example for $\delta = 2$ would be as follows,
\begin{align}
&[\textrm{Hansel } (\delta = 2)]: \nn
&{{|2\rangle \langle 3|}} \textit{Famous American foods } {|2\rangle}  \textit{created across United}  \nn 
&\textit{ States. Connecticut diner claims creation of the }{|1\rangle}\textit{ham-}  \nn
&\textit{burger. Onion rings were courtesy of cook at Pig Stand } \nn
&{|0\rangle} \textit{in Texas.} \nonumber 
\end{align}
We randomly choose $20\%$ of the samples and assign $l-1, l-2, \cdots, l-\delta$ uniformly. 
We also label mask $N = 10$ tokens preceding $| 0\rangle$ while training. 
This is to ensure that the model does not explicitely learn incomplete sentences preceding $| 0\rangle$.
A nonzero $\delta$ plays a crucial role in the Hansel framework preventing abrupt termination and its treatment while training will be detailed in the following section.

\subsection{The Hansel framework}

The Hansel framework essentially finetunes the pre-trained model with the Hansel dataset. 
One full example consists of a source, a prompt including the target length of the output, and the Hansel augmented output. 
The target length information is provided as the first special token.

We compare the Hansel framework with two baseline methods -- prompt finetuning and the finetuning without length information.
We dub the two as the Gretel and vanilla framework, respectively.
Comparing the three frameworks' training example, Hansel has [source]-[prompt with target length (special token)]-[Hansel output]; Gretel has [source]-[prompt with target length]-[original output]; and vanilla has [source]-[prompt without target length]-[original output].
For inference Gretel have [source]-[prompt with target length] as the context, while Hansel has [source]-[prompt with target length (special token)].
Vanilla has two versions, [source]-[prompt with/without target length]. 
We specifically dub the former (with target length) version as vanilla$^*$.

An analogy with the three methods can be made for a person giving a talk. 
The vanilla model would correspond to giving a talk for a specific duration without any time-keeping practice. 
Gretel is trained by knowing how long each practice talk was, and required to give one for a certain time. 
Hansel's practice talk was done with the knowledge of the total time, and a time-keeper who reminds the speaker every 5 minutes -- and the same time-keeper attends the actual talk as well. 
The schematic of the three frameworks is depicted in Figure~\ref{fig:hansel}.
We also present training and inference examples for each in the Appendix.

\section{Experimental Setup}

We used batch size 512 and AdamW optimizer~\cite{loshchilov2018decoupled} with $5\times 10^{-5}$ learning-rate and parameters $\beta_1=0.9$, $\beta_2=0.95$.
We finetune the pretrained LLMs for 2 epochs.
See the Appendix for how the number of epochs affects the performance. 
Finetuning and inference are conducted using 8 Nvidia Tesla V100 GPUs (40GB).
We set the maximum token number to 1722 and truncated longer examples.

\subsection{Dataset}

We consider two tasks in this work, text summarization and dialogue response.
Experiments are done with two datasets for each task.

\paragraph{CNN/DM}~\cite{cnndm} is a large-scale news dataset from the Cable News Network (CNN) and the Daily Mail (DM). We follow \cite{nallapati-etal-2016-abstractive}, and treat the news articles as the source documents and the corresponding highlights (usually in bullet points) as the summaries. 

\paragraph{XSum}~\cite{xsum} is a highly abstractive summarization dataset of news articles from the British Broadcasting Corporation (BBC).
BBC has one-line headlines at the beginning of each article, which are regarded as summaries. 

\paragraph{DailyDialog}~\cite{li2017dailydialog} is a multi-turn dialogue dataset on various daily conversation topics, including relationships, ordinary life, and work. 
The raw data are crawled from English-learning websites and thus are human-written.

\paragraph{MultiWOZ}~\cite{zang2020multiwoz} is a task-oriented dialogue dataset spanning 8 domains -- restaurant, hotel, attraction, taxi, train, hospital, bus, and police.
MultiWOZ also consists of human-human written conversation.

\subsection{Metric}

For evaluation, we use both metrics that measure the performance of the length control and the quality of the output.

\paragraph{Mean Absolute Error} To evaluate the length control ability of the model, we simply use the mean absolute error (MAE) of the output length from the target length. 
The MAE is measured as follows:
\begin{align}
    \textrm{MAE} = \frac{1}{N} \sum^N \left| l(\textrm{output}) - l(\textrm{target}) \right|.
\end{align}
$l(x)$ is the length of $x$ and $N$ is the number of samples. 

\paragraph{ROUGE} Recall-Oriented Understudy for Gisting Evaluation \cite{rouge} measures the lexical similarity between two texts by comparing the overlap of $n$-grams. 
Variants of ROUGE include the $n$-gram matching ROUGE-$n$ (R-$n$) and the longest common subsequence matching ROUGE-L (R-L). 
Among these, we show R-L in the main text and include results for R-1, R-2 in the Appendix. 
We use the Google Research implementation of ROUGE\footnote{https://github.com/google-research/google-research/tree/master/rouge} with the default settings.

\paragraph{G-Eval} To overcome the lexical bias from ROUGE, we also evaluate G-Eval~\cite{geval}, which is a framework for using LLMs with chain-of-thoughts (CoT) and a form-filling paradigm to assess the quality of natural language generation outputs. 
For the summarization task, the evaluation is in coherence (on a scale of 1-5), consistency (1-5), fluency (1-3), relevance (1-5); for the dialogue task, naturalness (1-5), coherence (1-5), engagingness (1-3), and groundedness (1-5). 
The score scale and the used prompts are from the original paper~\cite{geval}.
We use GPT-4~\cite{achiam2023gpt} as the evaluator.

\begin{table}[t]
    \centering
    \begin{tabular}{ccccc}
    \Xhline{3\arrayrulewidth} \\[-2.ex]
    Dataset & Model & MAE & R-L & G-Eval \\
    \Xhline{3\arrayrulewidth} \\[-2.ex]
    \multirow{4}{*}{DailyDialog} & vanilla$^{*}$ & 7.67 & 0.11 & 2.54 \\ 
    & vanilla & 7.01 & 0.25 & 3.51 \\ 
    & Gretel & 0.24 & 0.27 & 3.41 \\
    & Hansel & {\textbf{0.09}} & \textbf{0.33} & \textbf{3.55} \\ \hline \\[-2.ex]
    \multirow{4}{*}{MultiWOZ} & vanilla$^{*}$ & 7.24 & 0.13 & 1.86 \\ 
 & vanilla & 5.50 & 0.30 & \textbf{3.59} \\ 
    & Gretel & 0.11 & \textbf{0.34} & 3.55 \\
    & Hansel & \textbf{0.05} & \textbf{0.34} & 3.54 \\ \hline \\[-2.ex]
    \multirow{4}{*}{XSum} & vanilla$^{*}$ & 5.35 & 0.33 & 3.18 \\ 
 & vanilla & 5.35 & 0.33 & 3.20 \\ 
    & Gretel & 0.47 & 0.33 & 3.22 \\
    & Hansel & \textbf{0.26} & \textbf{0.34} & \textbf{3.26} \\ \hline \\[-2.ex]
    \multirow{4}{*}{CNN/DM} & vanilla$^{*}$ & 15.23 & 0.30 & 3.84 \\ 
 & vanilla & 14.82 & 0.30 & \textbf{3.86} \\ 
    & Gretel & 1.56 & 0.30 & 3.82 \\
    & Hansel & \textbf{0.39} & \textbf{0.31} & \textbf{3.86} \\ 
    \Xhline{3\arrayrulewidth}
    \end{tabular}
    \caption{The length control experiment using Phi-2 as the pre-trained LLM. The target length is set to the reference's actual length. Hansel and Gretel both showed good results without performance degradation. }
    \label{tab:phi2}
\end{table}

\subsection{Baseline Model}

\setlength{\tabcolsep}{1mm}
\begin{table}[t]
    \centering
    \begin{tabular}{ccccccc}
    \Xhline{3\arrayrulewidth} \\[-2.ex]
    \multirow{2.2}{*}{Dataset} & \multirow{2.2}{*}{Model}  &\multicolumn{5}{c}{$l(\textrm{target})$} \\[-.3ex]
    \cmidrule(lr){3-7}
    & & 5 & 20 & 50 & 80 & 130  \\
    \Xhline{3\arrayrulewidth} \\[-2.ex]
    \multirow{3}{*}{DailyDialog} & vanilla & 5.17 & 13.62 & 43.11 & 73.06 & 122.99 \\ 
    & Gretel & 0.15 & \textbf{0.33} & 2.01 & 7.15 & 18.51 \\
    & Hansel & {\textbf{0.06}} & {0.34} & {\textbf{0.43}} & {\textbf{0.79}} & \textbf{2.07}  \\ \hline \\[-2.ex]
    \multirow{3}{*}{MultiWOZ} & vanilla & 5.76 & 9.73 & 39.39 & 69.39 & 119.39 \\ 
    & Gretel & 0.12 & 0.17 & 2.89 & 9.96 & 23.85 \\
    & Hansel & \textbf{0.03} & \textbf{0.12} & \textbf{0.32} & \textbf{0.78} & \textbf{7.47} \\ \hline \\[-2.ex]
    \multirow{3}{*}{XSum} & vanilla & 13.71 & 4.30 & 32.34 & 62.29 & 112.23\\ 
    & Gretel & 0.96 & 0.50 & 1.80 & 3.42 & 10.37 \\
    & Hansel & \textbf{0.49} & \textbf{0.31} & \textbf{0.45} & \textbf{0.64} & \textbf{1.02} \\ \hline \\[-2.ex]
    \multirow{3}{*}{CNN/DM} & vanilla & 41.39 & 26.55 & 16.28 & 37.21 & 85.66 \\ 
    & Gretel & 1.49 & 0.59 & 1.54 & 3.21 & 7.38 \\
    & Hansel & \textbf{0.37} & \textbf{0.52} & \textbf{0.60} & \textbf{0.72} & \textbf{1.00} \\ 
    \Xhline{3\arrayrulewidth}
    \end{tabular}
    \caption{The length extrapolation results (MAE) of Phi-2. This is also plotted in Figure~\ref{fig:extrapolation}.}
    \label{tab:extrapolate}
\end{table}
\setlength{\tabcolsep}{6.pt}

\begin{figure*}[t]
    \centering
    \includegraphics[width=0.4\textwidth]{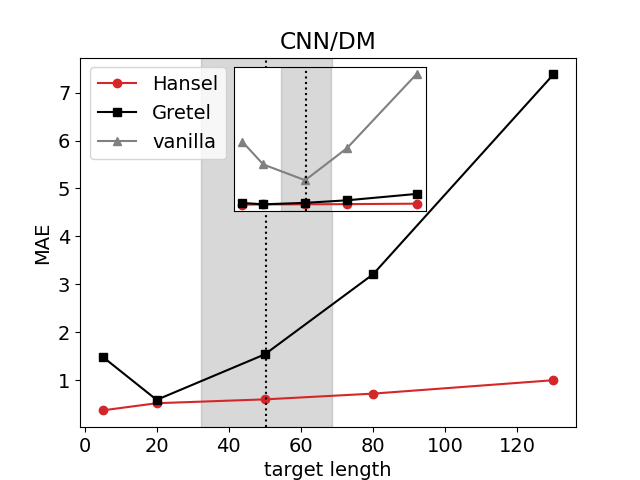}
    \includegraphics[width=0.4\textwidth]{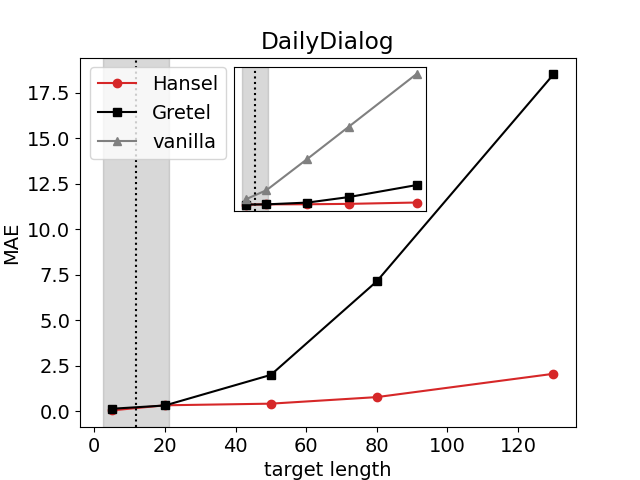}
    \caption{The extrapolation of the length control methods with different target lengths. The dashed line (shaded region) indicates the mean length ($\pm$ standard deviation) of the dataset. While the MAE of other methods increases drastically when the target length is different from that of the dataset, our method (Hansel) shows robust performance. } 
    \label{fig:extrapolation}
\end{figure*}

To demonstrate that Hansel is applicable regardless of the base model's positional representation method, we perform experiments on four different types of position methods -- Rotary positional embeddings, ALiBi, learned positional embeddings, and T5 bias. 
For consistency, we used models with parameter sizes within the range of 2.7B to 3.0B for all methods.

\paragraph{Rotary} Rotary positional embedding~\cite{su2024roformer} encodes the position with rotation matrices, only retaining information on purely relative position. 
For the model using rotary positional embedding, we use Microsoft's Phi-2 with 2.7B parameters~\cite{li2023textbooks}.

\paragraph{ALiBi} ALiBi~\cite{press2022train} is another relative positional method that is known to extrapolate well to longer sequences while inferencing. 
This adds a bias linear to the relative distance between the tokens.
We experiment on BLOOM-3B~\cite{workshop2022bloom} for ALiBi. 

\paragraph{Learned positional embedding} Positional embedding uses absolute embeddings that are added to token embeddings.
Unlike the other absolute method of positional encoding~\cite{vaswani2017attention}, the positional embeddings are learned throughout training. 
We use OPT-2.7B~\cite{zhang2022opt} for this method.

\paragraph{T5 Bias} The T5 bias~\cite{raffel2020exploring} is a relative positional embedding method~\cite{shaw2018self} similar to rotary and ALiBi but with a learned bias for each distance. 
We use the natural choice of T5-3B representing this method.

\setlength{\tabcolsep}{1mm}

\begin{table*}[t]
\centering
\begin{tabular}{cl}
\Xhline{3\arrayrulewidth} \\[-2.ex]
Task& \makecell[c]{~~~~~~~~~~~~~~~~~~~~~~~~~~~~~~~~Context} \\
\hline
\cellcolor{gray!15}Summarization &\cellcolor{gray!15}  \makecell[l]{Article: A group of scientists have invented a gadget which turns pee into electricity. They hope it could\\ be used in some poorer countries, as a way to create clean and renewable energy. The system was devel-\\oped by scientists at the Bristol Bioenergy Centre (BBiC) in the Bristol Robotics Laboratory at the Uni-\\versity of the West of England (UWE Bristol). The invention will be tried out this week on festival-goers\\ at the Glastonbury music festival in Somerset. Urine, or wee is fed into a special machine called a micro-\\bial fuel cell, which contains a type of bacteria. These tiny bacteria break down the wee and create elec-\\trons. These electrons are then forced through an electronic circuit, creating a flow of electricity. This\\ electricity can then be used to power all sorts of things, including your phone! [Summarize in \# words.]} \\
5 words & Scientists have invented a generator. \\[.3ex]
20 words & Scientists have come up with a new way to make electricity -- by using wee, or urine, as a power source. \\[.3ex]
80 words & \makecell[l]{Scientists have invented a machine which turns wee into electricity. It works by using bacteria to break\\ down the wee. This creates electrons which are forced through an electronic circuit. This electricity can\\ then be used to power all sorts of things, including your phone! Scientists hope it could be used in some\\ poorer countries, as a way to create clean and renewable energy. It will be tried out this week on festival-\\goers at the Glastonbury music festival in Somerset.} \\
\hline
\cellcolor{gray!15} \makecell[c]{Abrupt \\termination} \phantom{---} & \cellcolor{gray!15}\makecell[l]{A: I am looking for a hotel.
B: There are a total of 9 hotels. Huntingdon Marriotte Hotel is a good choice\phantom{-}\\ with a 4 star rating. Would you like there phone number?
A: I actually need a place that has free wifi\\ and is in the south part of town. And it needs to be a hotel, not a guesthouse. [Reply in 18 words.]}\\
$\delta = 0$ & \makecell[l]{There are two hotels that meet your needs. The Aylesbray Lodge Guest House and The Lensfield Hotel.\\ Any?}\\[.3ex]
$\delta = 1$ & The Lensfield Hotel is a hotel in the south with free wifi. Would you like to book it?\\
\hline
\cellcolor{gray!15}Multiple units & \cellcolor{gray!15} \makecell[l]{A: Have you any round-neck sweater? [Reply in \# sentences and \# words.]} \phantom{\quad}\phantom{--------------------------------,}\\
4 s 20 w & \makecell[l]{We have. Take a look at this one. It's made of cotton. It's very breathable and it's not expensive either.} \\[.3ex]
3 s 30 w & \makecell[l]{We have some nice round-neck sweaters. They are the latest fashion and are very popular with young\\ people. Why don't you try one on and see how you like it?} \\[.3ex]
2 s 40 w & \makecell[l]{We have some nice round-neck sweaters in different styles and colors. If you like this one, you can try\\ it on in the fitting room over there, and if you like it, you can buy it at the cashier's desk.} \\
\Xhline{3\arrayrulewidth}
\end{tabular}
\caption{Example context and outputs of the Phi-2 Hansel model. $\Delta = 20$ and $\delta = 1$ unless otherwise stated. }
\label{tab:example}
\end{table*}

\setlength{\tabcolsep}{6.pt}

\section{Results}
\label{sec:results}

\begin{table}[t]
    \centering
    \begin{tabular}{cccccc}
    \Xhline{3\arrayrulewidth} \\[-2.ex]
    \multirow{2.2}{*}{Dataset} & \multirow{2.2}{*}{$\Delta$} & \multicolumn{4}{c}{$\delta$} \\[-.3ex]
    \cmidrule(lr){3-6}
    & & 0 & 1 & 3 & 5 \\
    \Xhline{3\arrayrulewidth} \\[-2.ex]
    \multirow{3}{*}{DailyDialog} & 10 & 0.05 & {\textbf{0.08}} & {0.12} & {0.16} \\ 
    & 20 & 0.08 & {0.09} & {0.12} & {0.16} \\
    & 40 & 0.16 & {0.13} & {0.17} & {0.17} \\ \hline \\[-2.ex]
    \multirow{3}{*}{MultiWOZ} & 10 & 0.03 & \textbf{0.04} & 0.05 & 0.07 \\ 
    & 20 & 0.05 & 0.05 & 0.07 & 0.09 \\
    & 40 & 0.09 & 0.08 & 0.08 & 0.09 \\ \hline \\[-2.ex]
    \multirow{3}{*}{XSum} & 10 & 0.10 & \textbf{0.20} & 0.26 & 0.37 \\ 
    & 20 & 0.19 & 0.26 & 0.32 & 0.38 \\
    & 40 & 0.27 & 0.25 & 0.45 & 0.44 \\ \hline \\[-2.ex]
    \multirow{3}{*}{CNN/DM} & 10 & 0.09 & 0.40 & 0.60 & 0.83 \\ 
    & 20 & 0.17 & \textbf{0.39} & 0.60 & 0.76 \\
    & 40 & 0.53 & 0.49 & 0.81 & 0.88 \\ 
    \Xhline{3\arrayrulewidth}
    \end{tabular}
    \caption{The effect of hyperparameters $\Delta$ and $\delta$ on the length control ability. All entries are MAE compared with the reference length. Minimum values excluding $\delta = 0$ in bold. }
    \label{tab:hyperparam}
\end{table}

We first use Phi-2, which adapts rotary positional embedding, as the baseline model and finetune it with the four datasets in the three frameworks -- vanilla, Gretel, and Hansel. 
We then measure their performance using the test sets, asking the model to perform the task in a length-controlled manner. 
Here the target length was that of the reference. 
We set $\Delta = 20$ and $\delta = 1$ for this experiment.
Note that when the model infinitely generates repeated output, we exclude them from the statistics such as MAE, ROUGE-L, and G-Eval.
Instead we count the infinite generates samples separately and present them in the next section.

The result is shown in Table~\ref{tab:phi2}.
First, we comment on the vanilla$^*$. 
Vanilla$^*$ uses the prompt with target length while trained without such target. (recall that vanilla is the conventional fine-tuned model)
In dialogue datasets, this results in a model breakdown and the output quality greatly degrades, which can be checked in the ROUGE-L and G-Eval scores. 
Length control is meaningless if the model loses its generation ability, and therefore we do not consider the vanilla$^*$ model from now on. 
The vanilla model does not attempt the length control as the prompt does not include the target length, but its results are included in Table~\ref{tab:phi2} to demonstrate the other models (Hansel and Gretel) does not have quality degradation (ROUGE-L and G-Eval).
We include an expanded version of Table~\ref{tab:phi2} in the Appendix which includes R-1, R-2, and the four different measures in G-Eval. 

The Gretel model shows significant improvement over the vanilla on length control with a much smaller MAE. 
The Hansel model also shows good length control performance, better than that of Gretel. 
Hansel's MAE decrease over that of Gretel ranged from 45\% to 75\% among datasets. 
One salient feature is that Hansel's length control ability did not degrade as much for long-target datasets, while the MAE tends to be larger with a longer target length.
Due to its long average target length, CNN/DM has the largest MAE for all three frameworks, however Hansel improved the MAE by a significant 75\% over Gretel.

\paragraph{Target length extrapolation}

While the result in Table~\ref{tab:phi2} showed a better length control ability of Hansel, especially for long targets, one might consider Gretel model already has fairly small MAE and would be enough.
To demonstrate the true strength of Hansel, we now turn to a different and perhaps a more important setting of length control with arbitrary target length. 

There is a distinction between length control \emph{over a dataset} and the model having a \emph{general} length control capability regardless of the trained dataset. 
With the target length set as the reference length, Table~\ref{tab:phi2} demonstrated the former. 
For the model to claim general length control ability, it should be able to target arbitrary lengths, different from the reference length seen during training. 
We target 5, 20, 50, 80, and 130 words and measure the MAE for each case. 
The results are shown in Table~\ref{tab:extrapolate} and is plotted in  Figure~\ref{fig:extrapolation}. 
(the other two plots are presented in the Appendix)
We also show an example of the Hansel model outputs for a summarization task in Table~\ref{tab:example}, requested to generate outputs in various lengths. 

For all datasets and models, the MAE increases as the target length deviates from $\pm$(standard deviation) of the dataset (which is depicted as shaded regions in the figure).
However, Hansel shows a robustly small (order of magnitude smaller than Gretel for longest targets) MAE throughout the test range. 
This indicates that the Hansel model has learned how to control the output length in general, while Gretel's length control ability is more specific to the trained dataset.
This ability of Hansel is beneficial for both training efficiency and robustness in actual applications of length control.

\paragraph{The effect of hyperparameters}

We investigate the effect of the hyperparameters $\Delta$ and $\delta$. 
We vary $\Delta$ from $\{10, \,20, \,40\}$ and $\delta$ from $\{0, \, 1, \, 3, \, 5 \}$ and measure the MAE and show the results in Table~\ref{tab:hyperparam}.

The MAE increases with $\delta$ as we have trained the model with examples with larger residuals. 
Although $\delta = 0$ results were best in terms of MAE, we avoid selecting those from the quality perspective. 
Manually going through $\delta = 0$ outputs we identified instances where the output was abruptly terminated near the target length. 
We show an example of this in Table~\ref{tab:example}. 
The nonzero $\delta$ was introduced for precisely this reason and the result suggests $\delta = 1$ is sufficient for that purpose.
Small $\Delta$ is also unnecessary (as it generates excessive special tokens) and we conclude that $\Delta = 20$ and $\delta = 1$ is the best combination of hyperparameters. 

\begin{table}[t]
\centering
\begin{tabular}{ccccc}
\Xhline{3\arrayrulewidth} \\[-2.ex]
Dataset & Model & ALiBi & Learned & T5 Bias \\ 
\Xhline{3\arrayrulewidth} \\[-2.ex]
\multirow{3}{*}{DailyDialog} & vanilla & 6.98 & 6.00 & 8.69   \\
& Gretel & 0.24 & 0.60 & 0.33  \\ 
 & Hansel & {\textbf{0.15}} & {\textbf{0.21}} & {\textbf{0.18}}  \\ \hline\\[-2.ex]
\multirow{3}{*}{MultiWOZ} & vanilla & 5.67 & 5.68 & 5.63  \\
 & Gretel & 0.14 & 0.21 & 0.25 \\ 
 & Hansel & \textbf{0.12} & \textbf{0.11} & \textbf{0.22}  \\ \hline\\[-2.ex]
\multirow{3}{*}{XSum} & vanilla & 5.25 & 5.03 & 4.73 \\
& Gretel & 0.47 & 0.69 & 0.92  \\ 
& Hansel & \textbf{0.39} & \textbf{0.37} & \textbf{0.47}  \\ \hline\\[-2.ex]
\multirow{3}{*}{CNN/DM} & vanilla & 14.46 & 14.70 & 13.80  \\
& Gretel & 1.78 & 2.80 & 3.62 \\ 
& Hansel & \textbf{0.98} & \textbf{0.69} & \textbf{1.09} \\ 
\Xhline{3\arrayrulewidth}
\end{tabular}
\caption{The length control experiments showing MAE with baseline models using different positional embeddings. The target length is set to the reference's actual length.}
\label{tab:encodings}
\end{table}

\paragraph{Various positional embeddings}

As the Hansel framework only requires the Hansel dataset and does not alter the model architecture or positional embeddings, it can be applied to any baseline models.
To demonstrate this, we experiment on three more models, BLOOM, OPT, and T5. 
These represent models with ALiBi, learned, and T5 positional embeddings, respectively. 
Moreover, the T5 is an encoder-decoder model while the others are decoder models. 
The results are presented in Table~\ref{tab:encodings}.

While each model has a different base performance, the tendency is largely similar to Table~\ref{tab:phi2} -- Hansel controls length the best without significant loss of quality. 
This demonstrates the adaptability of the Hansel framework. 
\paragraph{Infinite generation}

Additionally, there is an interesting byproduct of this length control. Infinite generation is a quite common problem in LLMs, which gets more prominent for small on-device models. There are reports of this issue even in very recent models such as Llama 3~\cite{dubey2024llama}. If the model has the length controlling ability, this automatically solves the infinite generation issue. 
For the length extrapolation results in Table~\ref{tab:extrapolate}, vanilla and Gretel experienced infinite generation 8.4 and 15.3 times per 10,000 generations, respectively, while there was no such incident for Hansel.
Detailed data of this experiment is shown in the Appendix.

\paragraph{Versatility and generalizability}

The framework is also versatile enough to control multiple units. 
For example, it is possible to train a model that can control both the number of sentences and words. 
One just needs to prepare the Hansel dataset which consists of two different families of special tokens, each for a sentence and word.
An identical algorithm can be applied for augmentation. 
Examples of sentence and word number control are shown in Table~\ref{tab:example}.
Moreover, Hansel's length control can also work as on the fly -- one just need to specify the special token to turn on the length control ability.

We have also performed ablation studies to ensure Hansel does not affect the output distribution and is generalizable to other domains and tasks. 
The model's output on a separate task was not affected by the special tokens, and found that the generalization was possible with only 50 additional instruction samples.
Please refer to the Appendix for the details of the ablation studies.

\section{Conclusion}

We have introduced the Hansel framework that efficiently enables LLMs to control the length of their output.
The Hansel framework only utilizes the Hansel dataset which is an augmentation from the original dataset, and thus can be applied to models regardless of architecture and positional encoding.
Moreover, it is large pre-trained model friendly as it works through finetuning, and retains its length control ability even when the target length is out of the typical range of the finetuned dataset. 
The method outperforms the traditional prompt-based finetuning in the MAE from the target length. 
It especially showed significant improvement in extrapolated target length, demonstrating that the model learns the general means of length control.

\bibliography{Hansel_arXiv}

\begin{thebibliography}{45}
\providecommand{\natexlab}[1]{#1}

\bibitem[{Achiam et~al.(2023)Achiam, Adler, Agarwal, Ahmad, Akkaya, Aleman,
  Almeida, Altenschmidt, Altman, Anadkat et~al.}]{achiam2023gpt}
Achiam, J.; Adler, S.; Agarwal, S.; Ahmad, L.; Akkaya, I.; Aleman, F.~L.;
  Almeida, D.; Altenschmidt, J.; Altman, S.; Anadkat, S.; et~al. 2023.
\newblock Gpt-4 technical report.
\newblock \emph{arXiv preprint arXiv:2303.08774}.

\bibitem[{Anil et~al.(2023)Anil, Dai, Firat, Johnson, Lepikhin, Passos,
  Shakeri, Taropa, Bailey, Chen et~al.}]{anil2023palm}
Anil, R.; Dai, A.~M.; Firat, O.; Johnson, M.; Lepikhin, D.; Passos, A.;
  Shakeri, S.; Taropa, E.; Bailey, P.; Chen, Z.; et~al. 2023.
\newblock Palm 2 technical report.
\newblock \emph{arXiv preprint arXiv:2305.10403}.

\bibitem[{Arivazhagan et~al.(2019)Arivazhagan, Bapna, Firat, Lepikhin, Johnson,
  Krikun, Chen, Cao, Foster, Cherry et~al.}]{arivazhagan2019massively}
Arivazhagan, N.; Bapna, A.; Firat, O.; Lepikhin, D.; Johnson, M.; Krikun, M.;
  Chen, M.~X.; Cao, Y.; Foster, G.; Cherry, C.; et~al. 2019.
\newblock Massively multilingual neural machine translation in the wild:
  Findings and challenges.
\newblock \emph{arXiv preprint arXiv:1907.05019}.

\bibitem[{Dubey et~al.(2024)Dubey, Jauhri, Pandey, Kadian, Al-Dahle, Letman,
  Mathur, Schelten, Yang, Fan et~al.}]{dubey2024llama}
Dubey, A.; Jauhri, A.; Pandey, A.; Kadian, A.; Al-Dahle, A.; Letman, A.;
  Mathur, A.; Schelten, A.; Yang, A.; Fan, A.; et~al. 2024.
\newblock The llama 3 herd of models.
\newblock \emph{arXiv preprint arXiv:2407.21783}.

\bibitem[{Dubois et~al.(2024)Dubois, Li, Taori, Zhang, Gulrajani, Ba, Guestrin,
  Liang, and Hashimoto}]{dubois2024alpacafarm}
Dubois, Y.; Li, C.~X.; Taori, R.; Zhang, T.; Gulrajani, I.; Ba, J.; Guestrin,
  C.; Liang, P.~S.; and Hashimoto, T.~B. 2024.
\newblock Alpacafarm: A simulation framework for methods that learn from human
  feedback.
\newblock \emph{Advances in Neural Information Processing Systems}, 36.

\bibitem[{Fan, Grangier, and Auli(2018)}]{fan2018controllable}
Fan, A.; Grangier, D.; and Auli, M. 2018.
\newblock Controllable Abstractive Summarization.
\newblock In \emph{Proceedings of the 2nd Workshop on Neural Machine
  Translation and Generation}.

\bibitem[{Goyal, Li, and Durrett(2022)}]{goyal2022news}
Goyal, T.; Li, J.~J.; and Durrett, G. 2022.
\newblock News summarization and evaluation in the era of gpt-3.
\newblock \emph{arXiv preprint arXiv:2209.12356}.

\bibitem[{He et~al.(2022)He, Kryscinski, McCann, Rajani, and
  Xiong}]{he-etal-2022-ctrlsum}
He, J.; Kryscinski, W.; McCann, B.; Rajani, N.; and Xiong, C. 2022.
\newblock {CTRL}sum: Towards Generic Controllable Text Summarization.
\newblock In Goldberg, Y.; Kozareva, Z.; and Zhang, Y., eds., \emph{Proceedings
  of the 2022 Conference on Empirical Methods in Natural Language Processing}.

\bibitem[{Hermann et~al.(2015)Hermann, Kocisky, Grefenstette, Espeholt, Kay,
  Suleyman, and Blunsom}]{cnndm}
Hermann, K.~M.; Kocisky, T.; Grefenstette, E.; Espeholt, L.; Kay, W.; Suleyman,
  M.; and Blunsom, P. 2015.
\newblock Teaching machines to read and comprehend.
\newblock \emph{Advances in neural information processing systems}, 28.

\bibitem[{Jie et~al.(2023)Jie, Meng, Shang, Jiang, and Liu}]{jie2023prompt}
Jie, R.; Meng, X.; Shang, L.; Jiang, X.; and Liu, Q. 2023.
\newblock Prompt-Based Length Controlled Generation with Reinforcement
  Learning.
\newblock \emph{arXiv preprint arXiv:2308.12030}.

\bibitem[{Karpukhin et~al.(2020)Karpukhin, Oguz, Min, Lewis, Wu, Edunov, Chen,
  and Yih}]{karpukhin2020dense}
Karpukhin, V.; Oguz, B.; Min, S.; Lewis, P.; Wu, L.; Edunov, S.; Chen, D.; and
  Yih, W.-t. 2020.
\newblock Dense Passage Retrieval for Open-Domain Question Answering.
\newblock In \emph{Proceedings of the 2020 Conference on Empirical Methods in
  Natural Language Processing (EMNLP)}.

\bibitem[{Kikuchi et~al.(2016)Kikuchi, Neubig, Sasano, Takamura, and
  Okumura}]{kikuchi2016controlling}
Kikuchi, Y.; Neubig, G.; Sasano, R.; Takamura, H.; and Okumura, M. 2016.
\newblock Controlling Output Length in Neural Encoder-Decoders.
\newblock In \emph{Proceedings of the 2016 Conference on Empirical Methods in
  Natural Language Processing}.

\bibitem[{Kojima et~al.(2022)Kojima, Gu, Reid, Matsuo, and
  Iwasawa}]{kojima2022large}
Kojima, T.; Gu, S.~S.; Reid, M.; Matsuo, Y.; and Iwasawa, Y. 2022.
\newblock Large language models are zero-shot reasoners.
\newblock \emph{Advances in neural information processing systems}, 35.

\bibitem[{Lewis et~al.(2020)Lewis, Liu, Goyal, Ghazvininejad, Mohamed, Levy,
  Stoyanov, and Zettlemoyer}]{lewis2020bart}
Lewis, M.; Liu, Y.; Goyal, N.; Ghazvininejad, M.; Mohamed, A.; Levy, O.;
  Stoyanov, V.; and Zettlemoyer, L. 2020.
\newblock BART: Denoising Sequence-to-Sequence Pre-training for Natural
  Language Generation, Translation, and Comprehension.
\newblock In \emph{Proceedings of the 58th Annual Meeting of the Association
  for Computational Linguistics}.

\bibitem[{Li et~al.(2023)Li, Bubeck, Eldan, Del~Giorno, Gunasekar, and
  Lee}]{li2023textbooks}
Li, Y.; Bubeck, S.; Eldan, R.; Del~Giorno, A.; Gunasekar, S.; and Lee, Y.~T.
  2023.
\newblock Textbooks are all you need ii: phi-1.5 technical report.
\newblock \emph{arXiv preprint arXiv:2309.05463}.

\bibitem[{Li et~al.(2017)Li, Su, Shen, Li, Cao, and Niu}]{li2017dailydialog}
Li, Y.; Su, H.; Shen, X.; Li, W.; Cao, Z.; and Niu, S. 2017.
\newblock DailyDialog: A Manually Labelled Multi-turn Dialogue Dataset.
\newblock In \emph{Proceedings of the Eighth International Joint Conference on
  Natural Language Processing (Volume 1: Long Papers)}.

\bibitem[{Lin(2004)}]{rouge}
Lin, C.-Y. 2004.
\newblock {ROUGE}: A Package for Automatic Evaluation of Summaries.
\newblock In \emph{Text Summarization Branches Out}.

\bibitem[{Liu et~al.(2023{\natexlab{a}})Liu, Yuan, Fu, Jiang, Hayashi, and
  Neubig}]{10.1145/3560815}
Liu, P.; Yuan, W.; Fu, J.; Jiang, Z.; Hayashi, H.; and Neubig, G.
  2023{\natexlab{a}}.
\newblock Pre-train, Prompt, and Predict: A Systematic Survey of Prompting
  Methods in Natural Language Processing.
\newblock \emph{ACM Comput. Surv.}, 55(9).

\bibitem[{Liu et~al.(2024)Liu, Bai, Han, Weng, Xu, Cao, Wang, and
  Cai}]{liu2024length}
Liu, W.; Bai, Y.; Han, C.; Weng, R.; Xu, J.; Cao, X.; Wang, J.; and Cai, X.
  2024.
\newblock Length Desensitization in Directed Preference Optimization.
\newblock \emph{arXiv preprint arXiv:2409.06411}.

\bibitem[{Liu et~al.(2023{\natexlab{b}})Liu, Iter, Xu, Wang, Xu, and
  Zhu}]{geval}
Liu, Y.; Iter, D.; Xu, Y.; Wang, S.; Xu, R.; and Zhu, C. 2023{\natexlab{b}}.
\newblock {G}-Eval: {NLG} Evaluation using Gpt-4 with Better Human Alignment.
\newblock In \emph{Proceedings of the 2023 Conference on Empirical Methods in
  Natural Language Processing}.

\bibitem[{Liu, Jia, and Zhu(2022)}]{liu2022length}
Liu, Y.; Jia, Q.; and Zhu, K. 2022.
\newblock Length control in abstractive summarization by pretraining
  information selection.
\newblock In \emph{Proceedings of the 60th Annual Meeting of the Association
  for Computational Linguistics (Volume 1: Long Papers)}.

\bibitem[{Loshchilov and Hutter(2018)}]{loshchilov2018decoupled}
Loshchilov, I.; and Hutter, F. 2018.
\newblock Decoupled Weight Decay Regularization.
\newblock In \emph{International Conference on Learning Representations}.

\bibitem[{Lu et~al.(2024)Lu, Li, An, Zhao, He, Yin, and
  Sun}]{lu2024eliminating}
Lu, J.; Li, J.; An, S.; Zhao, M.; He, Y.; Yin, D.; and Sun, X. 2024.
\newblock Eliminating Biased Length Reliance of Direct Preference Optimization
  via Down-Sampled KL Divergence.
\newblock \emph{arXiv preprint arXiv:2406.10957}.

\bibitem[{Meng, Xia, and Chen(2024)}]{meng2024simpo}
Meng, Y.; Xia, M.; and Chen, D. 2024.
\newblock Simpo: Simple preference optimization with a reference-free reward.
\newblock \emph{arXiv preprint arXiv:2405.14734}.

\bibitem[{Nallapati et~al.(2016)Nallapati, Zhou, dos Santos,
  Gu\.{u}l{\c{c}}ehre, and Xiang}]{nallapati-etal-2016-abstractive}
Nallapati, R.; Zhou, B.; dos Santos, C.; Gu\.{u}l{\c{c}}ehre, {\c{C}}.; and
  Xiang, B. 2016.
\newblock Abstractive Text Summarization using Sequence-to-sequence {RNN}s and
  Beyond.
\newblock In Riezler, S.; and Goldberg, Y., eds., \emph{Proceedings of the 20th
  {SIGNLL} Conference on Computational Natural Language Learning}.

\bibitem[{Narayan, Cohen, and Lapata(2018)}]{xsum}
Narayan, S.; Cohen, S.~B.; and Lapata, M. 2018.
\newblock Don{'}t Give Me the Details, Just the Summary! Topic-Aware
  Convolutional Neural Networks for Extreme Summarization.
\newblock In Riloff, E.; Chiang, D.; Hockenmaier, J.; and Tsujii, J., eds.,
  \emph{Proceedings of the 2018 Conference on Empirical Methods in Natural
  Language Processing}.

\bibitem[{Park et~al.(2024)Park, Rafailov, Ermon, and
  Finn}]{park2024disentangling}
Park, R.; Rafailov, R.; Ermon, S.; and Finn, C. 2024.
\newblock Disentangling length from quality in direct preference optimization.
\newblock \emph{arXiv preprint arXiv:2403.19159}.

\bibitem[{Press, Smith, and Lewis(2022)}]{press2022train}
Press, O.; Smith, N.~A.; and Lewis, M. 2022.
\newblock Train short, test long: Attention with linear biases enables input
  length extrapolation.
\newblock In \emph{10th International Conference on Learning Representations
  (ICLR 2022)}.

\bibitem[{Rafailov et~al.(2024)Rafailov, Sharma, Mitchell, Manning, Ermon, and
  Finn}]{rafailov2024direct}
Rafailov, R.; Sharma, A.; Mitchell, E.; Manning, C.~D.; Ermon, S.; and Finn, C.
  2024.
\newblock Direct preference optimization: Your language model is secretly a
  reward model.
\newblock \emph{Advances in Neural Information Processing Systems}, 36.

\bibitem[{Raffel et~al.(2020)Raffel, Shazeer, Roberts, Lee, Narang, Matena,
  Zhou, Li, and Liu}]{raffel2020exploring}
Raffel, C.; Shazeer, N.; Roberts, A.; Lee, K.; Narang, S.; Matena, M.; Zhou,
  Y.; Li, W.; and Liu, P.~J. 2020.
\newblock Exploring the limits of transfer learning with a unified text-to-text
  transformer.
\newblock \emph{The Journal of Machine Learning Research}, 21(1): 5485--5551.

\bibitem[{Sanh et~al.(2022)Sanh, Webson, Raffel, Bach, Sutawika, Alyafeai,
  Chaffin, Stiegler, Raja, Dey, Bari, Xu, Thakker, Sharma, Szczechla, Kim,
  Chhablani, Nayak, Datta, Chang, Jiang, Wang, Manica, Shen, Yong, Pandey,
  Bawden, Wang, Neeraj, Rozen, Sharma, Santilli, Fevry, Fries, Teehan, Scao,
  Biderman, Gao, Wolf, and Rush}]{sanh2022multitask}
Sanh, V.; Webson, A.; Raffel, C.; Bach, S.; Sutawika, L.; Alyafeai, Z.;
  Chaffin, A.; Stiegler, A.; Raja, A.; Dey, M.; Bari, M.~S.; Xu, C.; Thakker,
  U.; Sharma, S.~S.; Szczechla, E.; Kim, T.; Chhablani, G.; Nayak, N.; Datta,
  D.; Chang, J.; Jiang, M. T.-J.; Wang, H.; Manica, M.; Shen, S.; Yong, Z.~X.;
  Pandey, H.; Bawden, R.; Wang, T.; Neeraj, T.; Rozen, J.; Sharma, A.;
  Santilli, A.; Fevry, T.; Fries, J.~A.; Teehan, R.; Scao, T.~L.; Biderman, S.;
  Gao, L.; Wolf, T.; and Rush, A.~M. 2022.
\newblock Multitask Prompted Training Enables Zero-Shot Task Generalization.
\newblock In \emph{International Conference on Learning Representations}.

\bibitem[{Shaw, Uszkoreit, and Vaswani(2018)}]{shaw2018self}
Shaw, P.; Uszkoreit, J.; and Vaswani, A. 2018.
\newblock Self-Attention with Relative Position Representations.
\newblock In \emph{Proceedings of the 2018 Conference of the North American
  Chapter of the Association for Computational Linguistics: Human Language
  Technologies, Volume 2 (Short Papers)}.

\bibitem[{Singhal et~al.(2024)Singhal, Goyal, Xu, and Durrett}]{singhal2024a}
Singhal, P.; Goyal, T.; Xu, J.; and Durrett, G. 2024.
\newblock A Long Way to Go: Investigating Length Correlations in {RLHF}.
\newblock In \emph{First Conference on Language Modeling}.

\bibitem[{Su et~al.(2024)Su, Ahmed, Lu, Pan, Bo, and Liu}]{su2024roformer}
Su, J.; Ahmed, M.; Lu, Y.; Pan, S.; Bo, W.; and Liu, Y. 2024.
\newblock Roformer: Enhanced transformer with rotary position embedding.
\newblock \emph{Neurocomputing}, 568: 127063.

\bibitem[{Sutskever, Vinyals, and Le(2014)}]{sutskever2014sequence}
Sutskever, I.; Vinyals, O.; and Le, Q.~V. 2014.
\newblock Sequence to sequence learning with neural networks.
\newblock \emph{Advances in neural information processing systems}, 27.

\bibitem[{Takase and Okazaki(2019)}]{takase-okazaki-2019-positional}
Takase, S.; and Okazaki, N. 2019.
\newblock Positional Encoding to Control Output Sequence Length.
\newblock In Burstein, J.; Doran, C.; and Solorio, T., eds., \emph{Proceedings
  of the 2019 Conference of the North {A}merican Chapter of the Association for
  Computational Linguistics: Human Language Technologies, Volume 1 (Long and
  Short Papers)}.

\bibitem[{Touvron et~al.(2023)Touvron, Martin, Stone, Albert, Almahairi,
  Babaei, Bashlykov, Batra, Bhargava, Bhosale et~al.}]{touvron2023llama}
Touvron, H.; Martin, L.; Stone, K.; Albert, P.; Almahairi, A.; Babaei, Y.;
  Bashlykov, N.; Batra, S.; Bhargava, P.; Bhosale, S.; et~al. 2023.
\newblock Llama 2: Open foundation and fine-tuned chat models.
\newblock \emph{arXiv preprint arXiv:2307.09288}.

\bibitem[{Vaswani et~al.(2017)Vaswani, Shazeer, Parmar, Uszkoreit, Jones,
  Gomez, Kaiser, and Polosukhin}]{vaswani2017attention}
Vaswani, A.; Shazeer, N.; Parmar, N.; Uszkoreit, J.; Jones, L.; Gomez, A.~N.;
  Kaiser, {\L}.; and Polosukhin, I. 2017.
\newblock Attention is all you need.
\newblock \emph{Advances in neural information processing systems}, 30.

\bibitem[{Workshop et~al.(2022)Workshop, Scao, Fan, Akiki, Pavlick, Ili{\'c},
  Hesslow, Castagn{\'e}, Luccioni, Yvon et~al.}]{workshop2022bloom}
Workshop, B.; Scao, T.~L.; Fan, A.; Akiki, C.; Pavlick, E.; Ili{\'c}, S.;
  Hesslow, D.; Castagn{\'e}, R.; Luccioni, A.~S.; Yvon, F.; et~al. 2022.
\newblock Bloom: A 176b-parameter open-access multilingual language model.
\newblock \emph{arXiv preprint arXiv:2211.05100}.

\bibitem[{Yu et~al.(2021)Yu, Wu, Zheng, XuanYuan, Fong, and
  Su}]{yu-etal-2021-lenatten}
Yu, Z.; Wu, Z.; Zheng, H.; XuanYuan, Z.; Fong, J.; and Su, W. 2021.
\newblock {L}en{A}tten: An Effective Length Controlling Unit For Text
  Summarization.
\newblock In Zong, C.; Xia, F.; Li, W.; and Navigli, R., eds., \emph{Findings
  of the Association for Computational Linguistics: ACL-IJCNLP 2021}.

\bibitem[{Yuan et~al.(2024)Yuan, Kulikov, Yu, Cho, Sukhbaatar, Weston, and
  Xu}]{yuan2024following}
Yuan, W.; Kulikov, I.; Yu, P.; Cho, K.; Sukhbaatar, S.; Weston, J.; and Xu, J.
  2024.
\newblock Following length constraints in instructions.
\newblock \emph{arXiv preprint arXiv:2406.17744}.

\bibitem[{Zang et~al.(2020)Zang, Rastogi, Sunkara, Gupta, Zhang, and
  Chen}]{zang2020multiwoz}
Zang, X.; Rastogi, A.; Sunkara, S.; Gupta, R.; Zhang, J.; and Chen, J. 2020.
\newblock MultiWOZ 2.2: A Dialogue Dataset with Additional Annotation
  Corrections and State Tracking Baselines.
\newblock In \emph{Proceedings of the 2nd Workshop on Natural Language
  Processing for Conversational AI}.

\bibitem[{Zhang et~al.(2020)Zhang, Zhao, Saleh, and Liu}]{zhang2020pegasus}
Zhang, J.; Zhao, Y.; Saleh, M.; and Liu, P. 2020.
\newblock Pegasus: Pre-training with extracted gap-sentences for abstractive
  summarization.
\newblock In \emph{International Conference on Machine Learning}. PMLR.

\bibitem[{Zhang et~al.(2022{\natexlab{a}})Zhang, Roller, Goyal, Artetxe, Chen,
  Chen, Dewan, Diab, Li, Lin et~al.}]{zhang2022opt}
Zhang, S.; Roller, S.; Goyal, N.; Artetxe, M.; Chen, M.; Chen, S.; Dewan, C.;
  Diab, M.; Li, X.; Lin, X.~V.; et~al. 2022{\natexlab{a}}.
\newblock Opt: Open pre-trained transformer language models.
\newblock \emph{arXiv preprint arXiv:2205.01068}.

\bibitem[{Zhang et~al.(2022{\natexlab{b}})Zhang, Zhang, Wang, Chen, and
  Wei}]{zhang2022latent}
Zhang, Y.; Zhang, X.; Wang, X.; Chen, S.-q.; and Wei, F. 2022{\natexlab{b}}.
\newblock Latent prompt tuning for text summarization.
\newblock \emph{arXiv preprint arXiv:2211.01837}.

\end{thebibliography}

\newpage
\appendix

\section{Dataset Statistics}
\label{app:stat}

We show the statistics of the test sets of the datasets in Table~\ref{tab:dataset}. 
Note that the reference length distribution is positively skewed, especially for the dialogue datasets.

\begin{table}[h]
    \centering
    \begin{tabular}{cccccc}
    \Xhline{3\arrayrulewidth} \\[-2.0ex]
    Dataset & mean & std & max & min & count \\
    \Xhline{3\arrayrulewidth} \\[-2.0ex]
    DailyDialog & 11.73 & 9.38 & 179 & 1 & 6205\\ 
    MultiWOZ & 13.41 & 6.68 & 54 & 2 & 13692\\ 
    XSum & 21.28 & 4.81 & 39 & 8 & 10138\\ 
    CNN/DM & 50.45 & 18.04 & 126 & 12 & 11297\\ 
    \Xhline{3\arrayrulewidth}
    \end{tabular}
    \caption{The statistics of the test set's reference length in the units of words.}
    \label{tab:dataset}
\end{table}

\section{Training and Inference Examples}
\label{app:prompt}

Here we show the example prompts for both training and inference, in the three frameworks. 
We choose the same source document for all prompts. 

\paragraph{Vanilla} Below is a training example of the vanilla framework for summarization task.
This consists of the source, prompt, and output, where the prompt is simply ``Summarize.''

\begin{align}
    &[\textrm{Source}]: \nonumber \\
    &\textit{The orca (Orcinus orca), or killer whale, is a } \nonumber \\
    &\textit{toothed whale that is the largest member of the} \nonumber \\
    &\textit{oceanic dolphin family. It is the only extant } \nonumber \\
    &\textit{species in the genus Orcinus. Orcas are recog- } \nonumber \\
    &\textit{nizable by their black-and-white patterned body.} \nonumber \\
    &[\textrm{Prompt}]: \nonumber\\
    &\textit{Summarize. } \nonumber\\
    &[\textrm{Output}]: \nonumber \\
    &\textit{Orcas are the largest member of the oceanic} \nonumber\\
    &\textit{dolphin family. They are also known as killer} \nonumber \\
    &\textit{whales.}\nonumber
\end{align}

\begin{figure*}[t]
    \centering
    \includegraphics[width=0.45\textwidth]{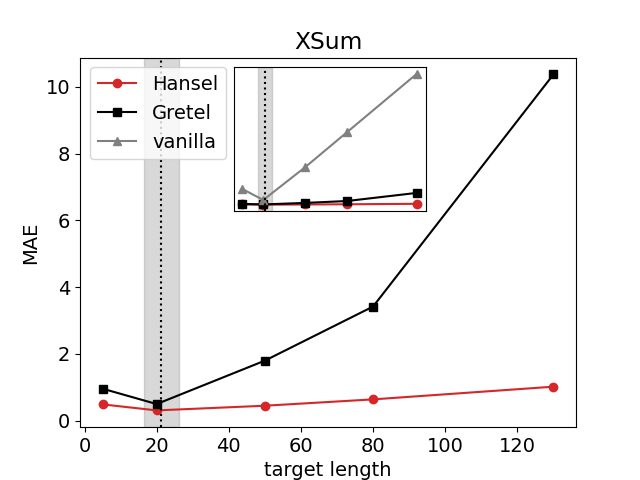}
    \includegraphics[width=0.45\textwidth]{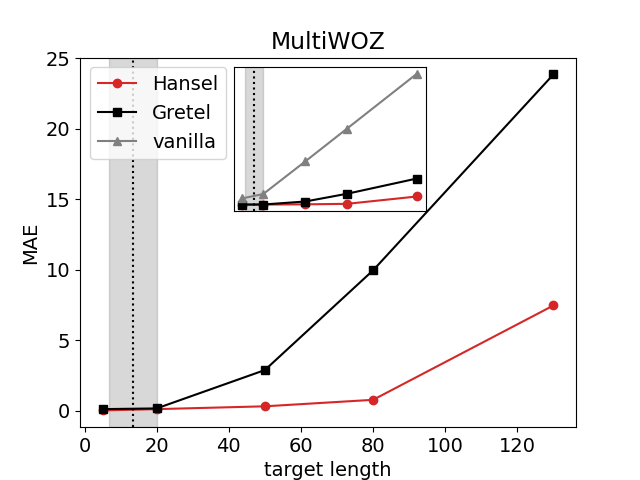}
    \caption{The extrapolation of the length control methods with different target lengths. The dashed line (shaded region) indicates the mean length ($\pm$ standard deviation) of the dataset. While the MAE increases drastically when the target length is different from that of the dataset, our method (Hansel) shows robust performance. }
    \label{fig:extrapolation2}
\end{figure*}

\paragraph{Gretel} 
An example of the Gretel training example is as follows.
In contrast to the vanilla example, the Gretel examples include the target length in the prompt as ``Answer in \#\# words.''
\begin{align}
    &[\textrm{Source}]: \nonumber \\
    &\textit{The orca (Orcinus orca), or killer whale, is a } \nonumber \\
    &\textit{toothed whale that is the largest member of the} \nonumber \\
    &\textit{oceanic dolphin family. It is the only extant } \nonumber \\
    &\textit{species in the genus Orcinus. Orcas are recog- } \nonumber \\
    &\textit{nizable by their black-and-white patterned body.} \nonumber \\
    &[\textrm{Prompt}]: \nonumber\\
    &\textit{Summarize. Answer in 17 words.} \nonumber\\
    &[\textrm{Output}]: \nonumber \\
    &\textit{Orcas are the largest member of the oceanic} \nonumber\\
    &\textit{dolphin family. They are also known as killer} \nonumber \\
    &\textit{whales.}\nonumber 
\end{align}

\paragraph{Hansel} Below is a training example of Hansel for the summarization task.
$\Delta = 10$ and the residual $\delta = 1$.
Special tokens are augmented to the output. 
Note that the output starts with the special token indicating 16 words left, $|1\rangle \langle 6|$, instead of 17 words because of the residual 1.
\begin{align}
    &[\textrm{Source}]: \nonumber \\
    &\textit{The orca (Orcinus orca), or killer whale, is a } \nonumber \\
    &\textit{toothed whale that is the largest member of the} \nonumber \\
    &\textit{oceanic dolphin family. It is the only extant } \nonumber \\
    &\textit{species in the genus Orcinus. Orcas are recog- } \nonumber \\
    &\textit{nizable by their black-and-white patterned body.} \nonumber \\
    &[\textrm{Prompt}]: \nonumber\\
    &\textit{Summarize.}\nonumber \\
    &[\textrm{Output}]: \nonumber \\
    &\textrm{$|1\rangle \langle 6|$} \textit{Orcas are the largest member of }\textrm{$|1\rangle$}\textit{the} \nonumber\\
    & \textit{oceanic dolphin family. They are also known as} \nonumber \\
    &\textit{killer }\textrm{$|0\rangle$}\textit{whales.}\nonumber
\end{align}

\paragraph{Inference} Inferences, regardless of vanilla, Gretel, or Hansel, have the basic structure of source and prompt including the target length. 
For the identical source with a target length of 23 is as below. 
\begin{align}
    &[\textrm{Source}]: \nonumber \\
    &\textit{The orca (Orcinus orca), or killer whale, is a } \nonumber \\
    &\textit{toothed whale that is the largest member of the} \nonumber \\
    &\textit{oceanic dolphin family. It is the only extant } \nonumber \\
    &\textit{species in the genus Orcinus. Orcas are recog- } \nonumber \\
    &\textit{nizable by their black-and-white patterned body.} \nonumber \\
    &[\textrm{Prompt}]: \nonumber\\
    &\textit{Summarize. Answer in 23 words.} \nonumber\\
    &[\textrm{Output}]: \nonumber \\
    &\textrm{$|2\rangle \langle 3|$}\nonumber 
\end{align}
Note that this is an example \textit{input} for inference, and the actual output (the 23 words summary in the above example) would follow as the model output.
Vanilla and Gretel are without the special token and only the Hansel inference includes the special token $|2\rangle \langle 3|$.
This special token can be considered as a switch that indicates the inference would be in Hansel style.

Note that we include 20\% of the vanilla dataset for the Hansel and Gretel training.
And for the Hansel training, 20\% of the non-vanilla examples consist of the Gretel dataset.
While these are optional, we made this choice to turn on and off the length control feature, and for the Hansel model to learn the Gretel ability.

\begin{figure}[t]
    \centering
    \includegraphics[width=0.47\textwidth]{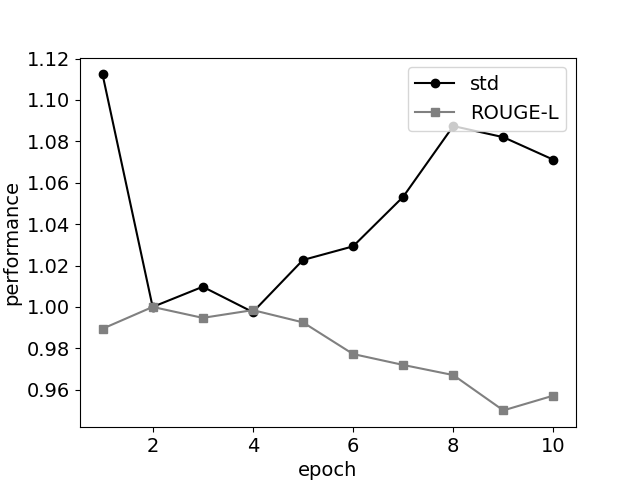}
    \caption{The MAE and ROUGE-L for the Phi-2 Hansel model as we increase the number of epochs. The performances are plotted as the ratio to the second epoch performance.}
    \label{fig:epoch}
\end{figure}
\begin{figure*}[t]
    \centering
    \includegraphics[width=0.75\textwidth]{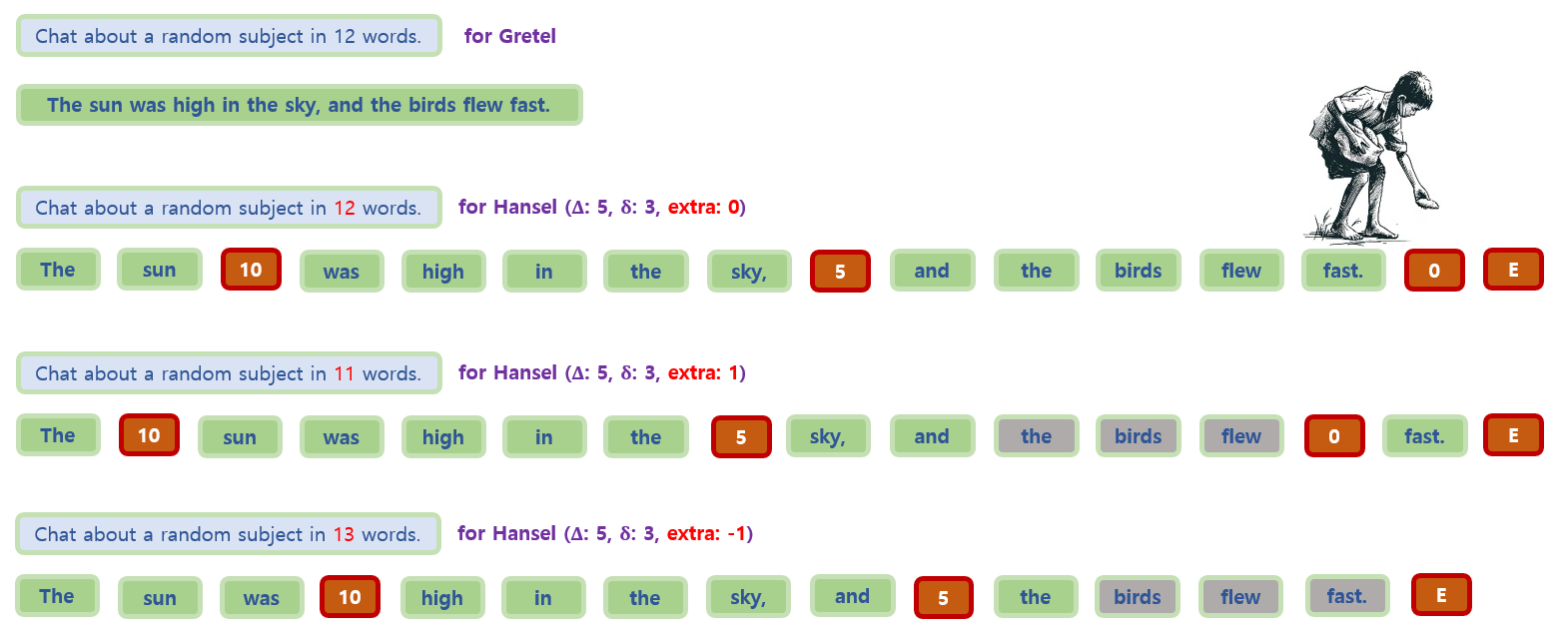}
    \caption{This figure explains how the special tokens are changed with different $\Delta$ and $\delta$. The grey tokens are label masked.}
    \label{fig:method}
\end{figure*}

\begin{table}[h]
    \centering
    \begin{tabular}{cccccc}
    \Xhline{3\arrayrulewidth} \\[-2.ex]
    \multirow{2.2}{*}{Dataset} & \multirow{2.2}{*}{$\Delta$} & \multicolumn{4}{c}{$\delta$} \\[-.3ex]
    \cmidrule(lr){3-6}
    & & 0 & 1 & 3 & 5 \\
    \Xhline{3\arrayrulewidth} \\[-2.ex]
    \multirow{3}{*}{DailyDialog} & 10 & 0.32 & {0.32} & {0.32} & {0.32} \\ 
    & 20 & 0.33 & {0.33} & {0.33} & {0.33} \\
    & 40 & 0.33 & {0.33} & {0.33} & {0.33} \\ \hline \\[-2.ex]
    \multirow{3}{*}{MultiWOZ} & 10 & 0.34 & 0.34 & 0.34 & 0.34 \\ 
    & 20 & 0.34 & 0.34 & 0.34 & 0.34 \\
    & 40 & 0.34 & 0.34 & 0.34 & 0.34 \\ \hline \\[-2.ex]
    \multirow{3}{*}{XSum} & 10 & 0.34 & 0.34 & 0.34 & 0.34  \\ 
    & 20 & 0.34 & 0.34 & 0.34 & 0.34  \\
    & 40 & 0.34 & 0.34 & 0.34 & 0.34  \\ \hline \\[-2.ex]
    \multirow{3}{*}{CNN/DM} & 10 & 0.33 & 0.33 & 0.33 & 0.33 \\ 
    & 20 & 0.34 & 0.34 & 0.34 & 0.34  \\
    & 40 & 0.34 & 0.34 & 0.34 & 0.34  \\ 
    \Xhline{3\arrayrulewidth}
    \end{tabular}
    \caption{The effect of hyperparameters $\Delta$ and $\delta$ on the length control ability. All entries are ROUGE-L score. }
    \label{tab:hyperparam-rouge}
\end{table}

\section{Effect of Training Epochs}
\label{app:epoch}

Here we show the effect of the number of training epochs in the Hansel framework. 
Figure~\ref{fig:epoch} shows the average standard deviation and the ROUGE-L score for all four datasets. 
Both the MAE and ROUGE-L show the best performance between epochs 2 to 4 and slowly degrade. 
The degradation is due to overfitting, which is common in finetuning tasks.
Based on this result we choose to train 2 epochs for all experiments in the main text.

\section{Effect of hyperparameters $\Delta$ and $\delta$}

In Table 4 of the main text, we have shown the effect of hyperparameters $\Delta$ and $\delta$ on the length control ability, using MAE as the metric. 
To justify the use of the best length-control model in Table 4, one should also confirm that the output quality is not affected by $\Delta$ and $\delta$. 
In Table~\ref{tab:hyperparam-rouge}, we show the ROUGE-L score of the same experiments in Table 4. 
It is clear from the results that $\Delta$ and $\delta$ does not affect the ROUGE-L score, and thus the output quality. 

We also re-demonstrate in detail of the role of hyperparameters $\Delta$ and $\delta$ in Figure~\ref{fig:method}.

\section{Ablation Studies}

Here we present two ablation studies to show that Hansel does not affect the output distribution and is generalizable to other domains and other tasks. 

To show the special token's affect on the output distribution, we observe whether the output of the Hansel model is different from either vanilla or Gretel on a separate (not Hansel finetuned) translation task. 
We prepare the vanilla, Gretel, and Hansel models with the Phi-3.5-mini (since Phi-2 only supports English we used the multi-lingual Phi-3.5 for this experiment). 
We finetuned the three models with the same WMT19 de-en dataset without any length information. 
The WMT19 test set BLEU (0.36) and COMET (0.85) scores were the same for all three models, and the sampled translation outputs were also almost identical. 
This supports that adding special tokens does not significantly change the output distribution for translation.

To demonstrate the generalization ability, we first finetuned two Phi-2 models with i) vanilla-summarization/Hansel-dialogue datasets and ii) vanilla-dialogue/Hansel-summarization datasets. 
These would not control the length of cross-tasks. 
However, by providing just 50 Hansel instruction samples, much smaller than the Hansel dataset used for initial finetuning, we reduced the MAE for summarization from 13.53 to 2.62 (model i) and dialogue from 31.50 to 0.93 (model ii) for both 40-word targets. 
Second, we prepared the Hansel model with Phi-3.5-mini and finetuned it with the WMT19 de-en dataset without length information (the same model used in the output distribution study). 
Similarly, we reduced the MAE of translation from 6.68 to 1.31 by 50 instruction samples (20-word target). 
These results suggest that Hansel is efficiently transferred among tasks and domains, and we expect the efficiency of transfer to increase as the finetuning tasks increase.
Eventually, we believe no additional tuning is required for a general instruction-tuned model. 
Note that the same experiment on the Gretel models was not successful.

\section{Additional Results}
\label{app:more_results}

We show the extrapolation experiment result similar to that of Figure 3 of main text, for the XSum and MultiWOZ dataset. 
This is shown in Figure~\ref{fig:extrapolation2}. 
The tendency of the three methods follows that of the CNN/DM and DailyDialog.

Table 1 and 5 of the main text only show ROUGE-L and the average G-Eval score for presentation. 
Here we elaborate on these performance metrics and show the ROUGE-1, ROUGE-2, and the four category scores of G-Eval. 
The result is in Table~\ref{tab:detail1} and Table~\ref{tab:detail2}.
\begin{table*}[t]
    \centering
    \begin{tabular}{ccccccccccc}
    \Xhline{3\arrayrulewidth} \\[-2.ex]
    Dataset & Baseline & Model & R-1 & R-2 & R-L & G-Eval & nat. & coh. & eng. & gr. \\
    \Xhline{3\arrayrulewidth} \\[-2.ex]
    \multirow{12}{*}{DailyDialog} & \multirow{3}{*}{Rotary} & vanilla  & 0.26 & 0.14 & 0.25 & 3.51 & 4.64 & 4.57 & 2.06 & 2.78 \\ 
    & & Gretel & 0.29 & 0.14 & 0.27 & 3.41 & 4.48 & 4.34 & 2.13 & 2.69 \\
    & & Hansel & 0.34 & 0.20 & 0.32 & 3.55 & 4.64 & 4.58 & 2.22 & 2.77 \\
    \cline{2-11}
     & \multirow{3}{*}{ALiBi} & vanilla & 0.31 & 0.19 & 0.29 & 3.34 & 4.43 & 4.26 & 2.03 & 2.63 \\ 
    && Gretel & 0.34 & 0.21 & 0.32 & 3.17 & 4.21 & 3.69 & 2.01 & 2.50 \\
    && Hansel & 0.33 & 0.20 & 0.32 & 3.32 & 4.32 & 4.15 & 2.16 & 2.64 \\
    \cline{2-11}
     & \multirow{3}{*}{Learned}& vanilla & 0.29 & 0.17 & 0.27 & 3.37 & 4.50 & 4.31 & 2.04 & 2.65 \\ 
    && Gretel & 0.32 & 0.18 & 0.30 & 3.43 & 4.50 & 4.37 & 2.14 & 2.69 \\
    && Hansel & 0.31 & 0.17 & 0.29 & 3.38 & 4.44 & 4.23 & 2.18 & 2.68 \\
    \cline{2-11}
     & \multirow{3}{*}{T5 Bias}& vanilla & 0.18 & 0.06 & 0.17 & 3.37 & 4.55 & 4.39 & 1.82 & 2.71 \\ 
    && Gretel & 0.23 & 0.08 & 0.22 & 3.36 & 4.40 & 4.30 & 2.04 & 2.69 \\
    && Hansel & 0.23 & 0.08 & 0.21 & 3.38 & 4.42 & 4.30 & 2.08 & 2.71 \\\hline \\[-2.ex]
    \multirow{12}{*}{MultiWOZ} & \multirow{3}{*}{Rotary} & vanilla & 0.33 & 0.16 & 0.30 & 3.59 & 4.77 & 4.65 & 2.11 & 2.82 \\ 
    && Gretel & 0.38 & 0.19 & 0.34 & 3.55 & 4.71 & 4.66 & 2.14 & 2.78 \\
    && Hansel & 0.38 & 0.19 & 0.34 & 3.54 & 4.70 & 4.55 & 2.13 & 2.77 \\ 
    \cline{2-11}
     &\multirow{3}{*}{ALiBi} &  vanilla & 0.30 & 0.14 & 0.27 & 3.50 & 4.68 & 4.51 & 2.07 & 2.76 \\
    && Gretel & 0.35 & 0.17 & 0.32 & 3.54 & 4.68 & 4.55 & 2.16 & 2.77 \\
    && Hansel & 0.35 & 0.17 & 0.32 & 3.54 & 4.68 & 4.57 & 2.13 & 2.79 \\
    \cline{2-11}
     & \multirow{3}{*}{Learned} & vanilla & 0.31 & 0.15 & 0.29 & 3.55 & 4.74 & 4.60 & 2.08 & 2.78 \\
    && Gretel & 0.37 & 0.19 & 0.34 & 3.57 & 4.73 & 4.61 & 2.15 & 2.80 \\
    && Hansel & 0.37 & 0.19 & 0.33 & 3.54 & 4.69 & 4.53 & 2.15 & 2.79 \\
    \cline{2-11}
     & \multirow{3}{*}{T5 Bias} & vanilla & 0.32 & 0.15 & 0.29 & 3.50 & 4.68 & 4.52 & 2.04 & 2.76 \\
    && Gretel & 0.38 & 0.19 & 0.34 & 3.55 & 4.71 & 4.54 & 2.17 & 2.76 \\
    && Hansel & 0.37 & 0.19 & 0.34 & 3.54 & 4.68 & 4.54 & 2.15 & 2.78 \\ 
    \Xhline{3\arrayrulewidth}
    \end{tabular}
    \caption{Additional dialogue dataset performance metrics ROUGE-1, ROUGE-2, and the four categories of G-Eval -- naturalness (scale of 1-5), coherence (1-5), engagingness (1-3), and groundedness (1-5). }
    \label{tab:detail1}
\end{table*}
\begin{table*}[t]
    \centering
    \begin{tabular}{ccccccccccc}
    \Xhline{3\arrayrulewidth} \\[-2.ex]
    Dataset & Baseline & Model & R-1 & R-2 & R-L & G-Eval & coh. & con. & flu. & rel. \\
    \Xhline{3\arrayrulewidth} \\[-2.ex]
    \multirow{12}{*}{XSum} & \multirow{3}{*}{Rotary} & vanilla  & 0.41 & 0.18 & 0.33 & 3.20 & 3.36 & 3.57 & 2.93 & 2.96 \\
    & & Gretel & 0.41 & 0.18 & 0.33 & 3.22 & 3.33 & 3.59 & 3.03 & 2.94 \\
    & & Hansel & 0.42 & 0.19 & 0.33 & 3.30 & 3.46 & 3.68 & 3.10 & 2.96 \\
    \cline{2-11}
     & \multirow{3}{*}{ALiBi} & vanilla & 0.39 & 0.17 & 0.31 & 3.01 & 3.01 & 3.34 & 2.74 & 2.94 \\
    && Gretel & 0.39 & 0.16 & 0.31 & 2.92 & 2.78 & 3.28 & 2.70 & 2.92 \\
    && Hansel & 0.39 & 0.16 & 0.31 & 2.96 & 2.87 & 3.31 & 2.75 & 2.91 \\
    \cline{2-11}
     & \multirow{3}{*}{Learned}& vanilla & 0.41 & 0.19 & 0.33 & 3.20 & 3.33 & 3.57 & 2.94 & 2.96 \\
    && Gretel & 0.41 & 0.19 & 0.33 & 3.16 & 3.18 & 3.55 & 2.97 & 2.96 \\
    && Hansel & 0.41 & 0.18 & 0.33 & 3.22 & 3.34 & 3.58 & 3.01 & 2.95 \\
    \cline{2-11}
     & \multirow{3}{*}{T5 Bias}& vanilla & 0.44 & 0.21 & 0.36 & 3.32 & 3.57 & 3.71 & 3.04 & 2.97 \\
    && Gretel & 0.44 & 0.21 & 0.36 & 3.37 & 3.59 & 3.75 & 3.17 & 2.96 \\
    && Hansel & 0.44 & 0.21 & 0.36 & 3.39 & 3.60 & 3.79 & 3.21 & 2.96 \\\hline \\[-2.ex]
    \multirow{12}{*}{CNN/DM} & \multirow{3}{*}{Rotary} & vanilla & 0.42 & 0.20 & 0.30 & 3.86 & 4.42 & 4.13 & 3.92 & 2.99 \\ 
    && Gretel & 0.43 & 0.20 & 0.30 & 3.82 & 4.40 & 4.04 & 3.84 & 2.98 \\
    && Hansel & 0.44 & 0.20 & 0.31 & 3.86 & 4.46 & 4.10 & 3.91 & 2.98 \\
    \cline{2-11}
     &\multirow{3}{*}{ALiBi} &  vanilla & 0.41 & 0.19 & 0.29 & 3.70 & 4.19 & 3.92 & 3.71 & 2.97 \\
    && Gretel & 0.42 & 0.19 & 0.29 & 3.67 & 4.15 & 3.90 & 3.69 & 2.56 \\
    && Hansel & 0.42 & 0.19 & 0.29 & 3.69 & 4.16 & 3.90 & 3.69 & 2.99 \\
    \cline{2-11}
     & \multirow{3}{*}{Learned} & vanilla & 0.41 & 0.20 & 0.30 & 3.82 & 4.37 & 4.08 & 3.84 & 2.99 \\
    && Gretel & 0.43 & 0.20 & 0.30 & 3.82 & 4.40 & 4.05 & 3.86 & 2.97 \\
    && Hansel & 0.43 & 0.20 & 0.30 & 3.81 & 4.38 & 4.03 & 3.85 & 2.99 \\
    \cline{2-11}
     & \multirow{3}{*}{T5 Bias} & vanilla & 0.43 & 0.21 & 0.31 & 3.91 & 4.58 & 4.15 & 3.92 & 2.99 \\
    && Gretel & 0.44 & 0.21 & 0.32 & 3.91 & 4.55 & 4.14 & 3.96 & 2.98 \\
    && Hansel & 0.44 & 0.21 & 0.32 & 3.91 & 4.55 & 4.13 & 3.96 & 2.99 \\
    \Xhline{3\arrayrulewidth}
    \end{tabular}
    \caption{Additional summarization performance metrics ROUGE-1, ROUGE-2, and the four categories of G-Eval -- coherence (scale of 1-5), consistency (1-5), fluency (1-3), and relevance (1-5). }
    \label{tab:detail2}
\end{table*}

\end{document}